\documentclass[10pt,twocolumn,letterpaper]{article}

\usepackage{iccv}
\usepackage{times}
\usepackage{epsfig}
\usepackage{graphicx}
\usepackage{caption}
\usepackage{stfloats}
\usepackage{amsmath}
\usepackage{booktabs}
\usepackage{amssymb}

\usepackage{bbm}
\usepackage{bm}

\def\ie{\emph{i.e}\onedot}

\usepackage[pagebackref=true,breaklinks=true,letterpaper=true,colorlinks,bookmarks=false]{hyperref}

\iccvfinalcopy 

\def\iccvPaperID{1872} 

\ificcvfinal\fi

\begin{document}

\title{
Unified Multi-Modal Latent Diffusion for \\Joint Subject and Text Conditional Image Generation
}


\author{Yiyang Ma\textsuperscript{1},
Huan Yang\textsuperscript{2},
Wenjing Wang\textsuperscript{1},
Jianlong Fu\textsuperscript{2}, and Jiaying Liu\textsuperscript{1}\\
\textsuperscript{1}Wangxuan Institute of Computer Technology, Peking University, \textsuperscript{2}Microsoft Research\\
\tt\small \textsuperscript{1}\{myy12769, daooshee, liujiaying\}@pku.edu.cn, \textsuperscript{2}\{huayan, jianf\}@microsoft.com
}



\twocolumn[{
\renewcommand\twocolumn[1][]{#1}
\maketitle
\begin{center}
    \captionsetup{type=figure}
    \includegraphics[width=\textwidth]{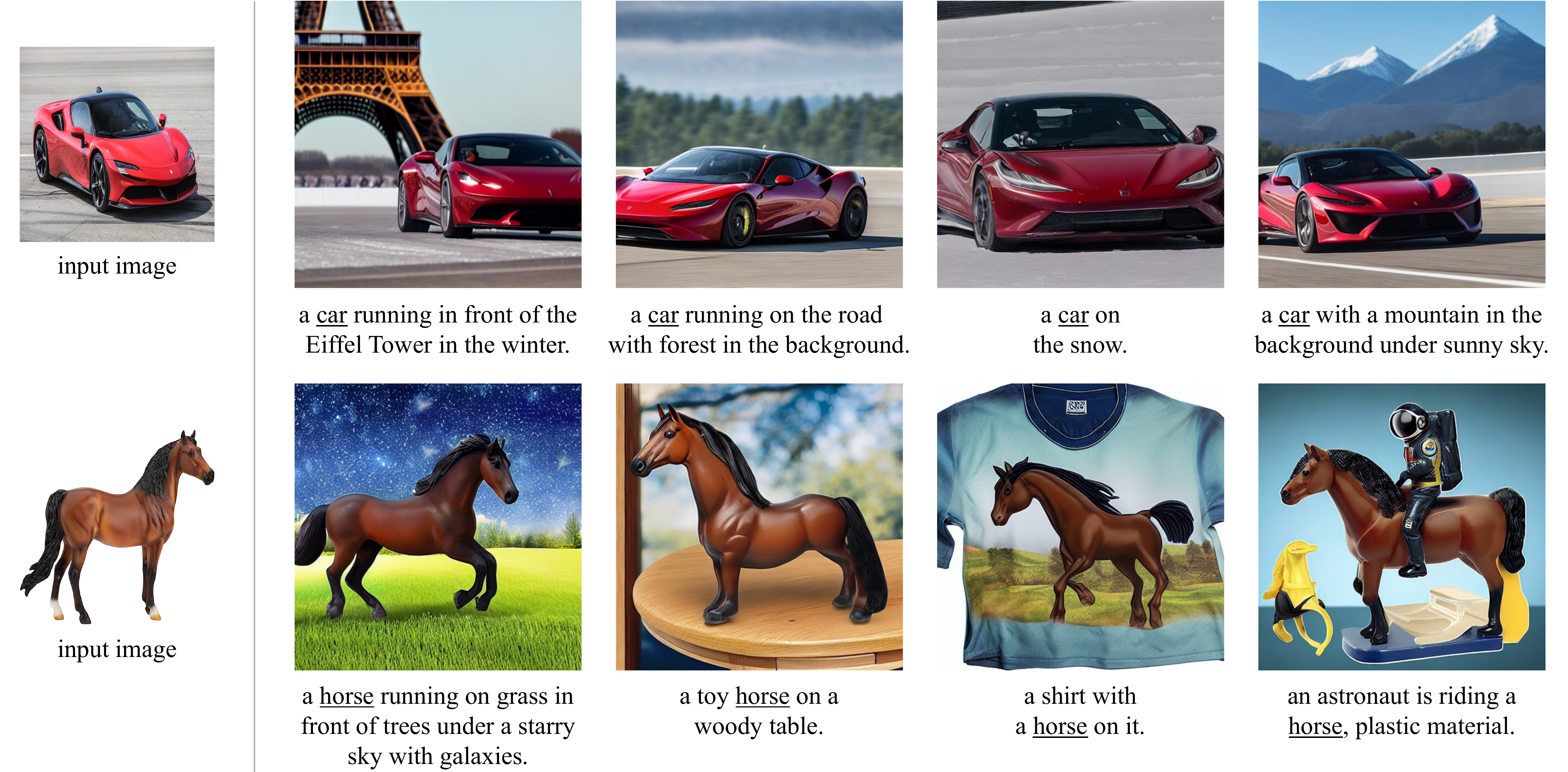}
    \captionof{figure}{
    Our approach can generate novel high-quality images by keeping the visual appearance of a subject provided in an image, and aligning the semantics of given texts. To achieve this goal, we encode the joint visual subject and text into a unified multi-modal latent space as guidance for image generation \textbf{without any finetuning}.
    }
    \label{fig: teaser}
\end{center}
}]

\ificcvfinal\thispagestyle{empty}\fi

\begin{abstract}

Language-guided image generation has achieved great success nowadays by using diffusion models. However, texts can be less detailed to describe highly-specific subjects such as a particular dog or a certain car, which makes pure text-to-image generation not accurate enough to satisfy user requirements. In this work, we present a novel \textbf{U}nified \textbf{M}ulti-\textbf{M}odal Latent \textbf{Diffusion} (\textbf{UMM-Diffusion}) which takes joint texts and images containing specified subjects as input sequences and generates customized images with the subjects. To be more specific, both input texts and images are encoded into one unified multi-modal latent space, in which the input images are learned to be projected to pseudo word embedding and can be further combined with text to guide image generation.
Besides, to eliminate the irrelevant parts of the input images such as background or illumination, we propose a novel sampling technique of diffusion models used by the image generator which fuses the results guided by multi-modal input and pure text input. By leveraging the large-scale pre-trained text-to-image generator and the designed image encoder, our method is able to generate high-quality images with complex semantics from both aspects of input texts and images.

\end{abstract}

\section{Introduction}

Just imagine that your favorite car appears all over the world or even in an oil painting. That is fascinating, right? However, synthesizing such images is not a simple task. The customized subjects you specify by giving images along with texts must be jointly processed and encoded into a unified latent space so that they can be used as guidance for image generation. The input data format is like ``A \underline{car} is running in front of the Eiffel Tower'' and an image of your car. That means we have to comprehend sequential images and texts jointly and use the ``comprehension'' as guidance to generate new images.


Existing text-to-image generation models cannot achieve such a goal, because they use only text encoders like \textit{T5} \cite{2020T5} as conditioning models, which can only encode pure text input. For multi-modal encoding, it is intuitive to think about cross-modal alignment encoders like \textit{CLIP} \cite{2021CLIP}. But they can only encode images and texts into their corresponding latent spaces in a separate way. It is nontrivial to build a conditioning model to encode multi-modal data into one unified latent space. Without this kind of conditioning model, directly building a multi-modality guided image generation model as we discussed before is quite difficult.

Recently, to avoid the difficulty of jointly encoding multi-modal data into one unified latent space, several methods of finding a special text token representing the specific input subject that can be processed by existing text-to-image models have been proposed \cite{2022TI, 2022DreamBooth}. Ruiz \etal \cite{2022DreamBooth} finetune the text-to-image model with few-shot data containing input images of a certain subject and texts containing a unique identifier followed by its class name. Gal \etal \cite{2022TI} use an inverse method to find a ``pseudo word'' indicating the input subject without optimizing the parameters of the text-to-image model. Nevertheless, these methods have to specially treat every input data which means great time and computing resource costs are needed and high-resolution images are also required (near or bigger than the generated images of the model) as training data. Their models are not general to directly take multi-modal data as input to guide the generation process but are specially tuned for each specific input. Different from them, our method takes joint texts and images in one sequence as input straightway and encodes them into one unified multi-modal latent space, which means the proposed method does not require such treatment on every single input. Meanwhile, our method can still get comparable results in most cases.

In this work, we define a new form of the joint subject and text conditional image generation process which takes unified texts and images in one sequence as guidance and propose \textbf{U}nified \textbf{M}ulti-\textbf{M}odal Latent \textbf{Diffusion} (\textbf{UMM-Diffusion}) to solve it. First, we give a specified task definition below. Given a caption, several images, and the positions of words in the caption which the images refer to, we manage to generate a new image described by the caption and containing subjects provided by the input images.
Then, in order to achieve the goal of such a joint subject and text conditional image generation, we present a joint text-and-image encoder as the conditioning model which encode the cross-modal input into one unified multi-modal latent space. The encoder first extracts the \textit{CLIP Image Embedding}s from the input images, then projects them to pseudo word embeddings, and last rearranges the pseudo word embeddings and the real word embeddings at the rest positions of the input text so that the combined word embedding sequence can be processed by \textit{CLIP Text Encoder}. The conditioning model leverages \textit{CLIP Encoder}s which could help to extract semantics from both images and texts.
Besides, to mitigate the problem of overfitting on the irrelevant part of input images like background or illumination, we propose a novel sampling technique of diffusion models which fuses the results of multi-modal guidance and pure text guidance. Several results are shown in Fig.~\ref{fig: teaser}.

In summary, our contributions are as follows:

\begin{itemize}
    \item We propose a new form of joint subject and text conditional image generation which takes texts and images of certain words in the texts in one sequence as input performing unified condition.
    \item We design a unified text-and-image encoder to encode mixed sequences of both texts and images into one multi-modal latent space so that they can be used as guidance for the image generation process.
    \item We present a novel sampling technique of diffusion models to mitigate the overfitting problem of the irrelevant part of input images which fuses multi-modal guided results and pure text guided results in one single denoising step.
\end{itemize}

\begin{figure*}[ht]
    \centering
    \includegraphics[width=\textwidth]{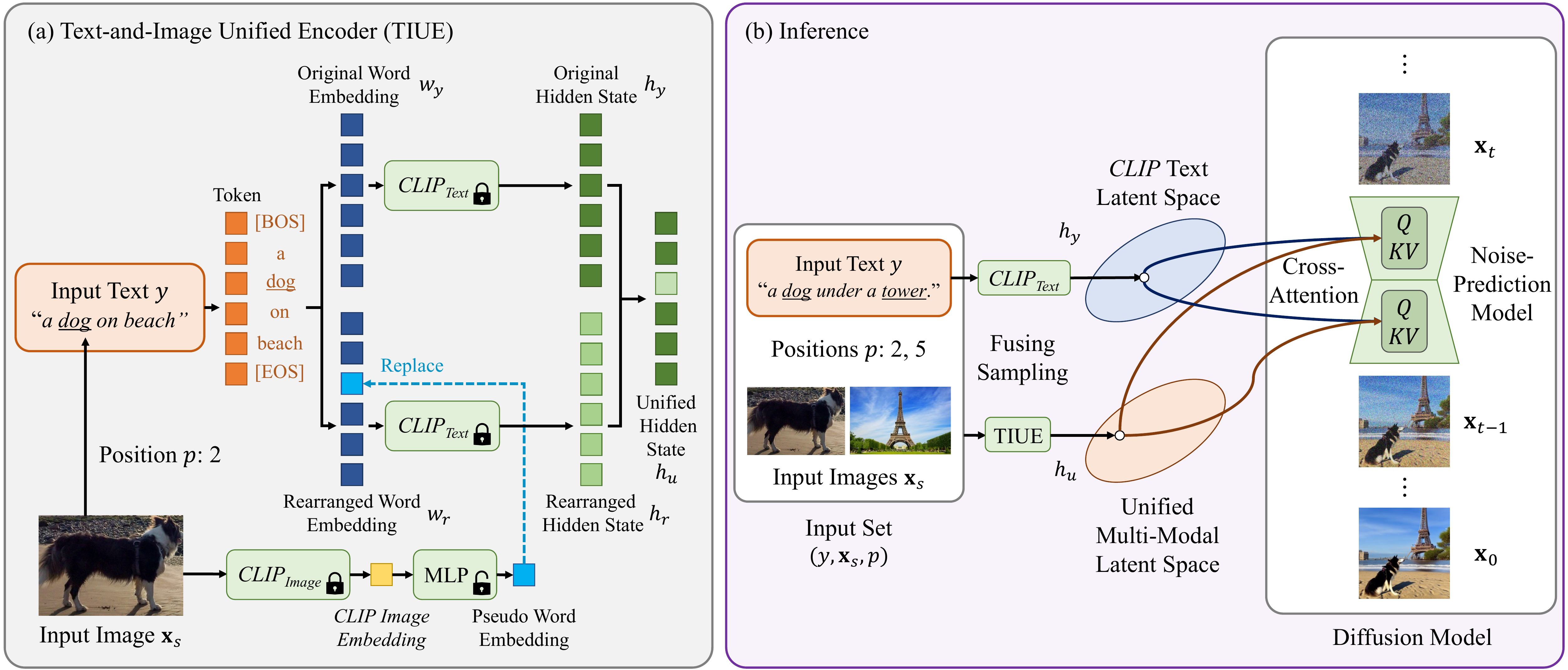}
    \caption{The approach of \textbf{UMM-Diffusion}. In TIUE, only the MLP which maps \textit{CLIP Image Embedding}s to pseudo word embeddings is trainable. The TIUE encodes joint texts and images to $h_u$ within the unified multi-modal latent space. During inference, both pure-text $h_y$ and multi-modal $h_u$ are extracted and used as guidance of generation process via the fusing sampling technique proposed in Sec. \ref{subsec: fuse sample}. For better visualization, we show a vanilla diffusion model in the figure while we use Stable Diffusion \cite{2022StableDiffusion} in practice.}
    \label{fig: method}
\end{figure*}

\section{Related Work}

\subsection{Compositing Subjects into Scenes}

Extracting a certain subject from a given image and compositing it into a new scene ensuring the integrity and reality of the generated image is a basic and very common scenario~\cite{2006Color, 2020Dovenet, 2020Improving, 2006Drag, 2018ST, 2003Poisson, 2001Color, 2010MultiScale, 2017Deep, 2019GP} of image editing.
Traditional methods~\cite{2006Color, 2006Drag, 2003Poisson, 2001Color, 2010MultiScale} manage to use statistic features of images to align the subjects and the new scenes.
Thanks to the development of deep learning, recent methods replacing statistic features with deep features achieve better performance~\cite{2020Dovenet, 2020Improving, 2018ST, 2017Deep, 2019GP}.
However, these methods assume that the subject and the scene are already harmonious.
As a result, they cannot create new views and are not tolerant of any unaligned structures of subjects and scenes, which limits the scope of application.
More recent methods leveraging 3D reconstruction~\cite{2021Nerf, 2021mipNerf} can generate new views of a certain object but require a lot of different view images of the object.
In short, synthesizing novel views of the given subjects with a limited number of images is quite non-trivial.
Compositing subjects into scenes is a challenging task.

\subsection{Text Conditional Image Generation}

Generating semantically aligned images conditioned by texts has drawn great attention.
Early methods~\cite{2016GAN-INT-CLS, 2017StackGAN, 2017SISGAN, 2018AttnGAN, 2018HD-GAN, 2018StackGAN++, 2019ControlGAN, 2019DM-GAN, 2019MirrorGAN, 2020DF-GAN} are mainly based on generative adversarial networks (GANs)~\cite{2014GAN, 2014cGAN} and trained on limited data.
To solve the problem of getting semantics from open-world texts and leverage generative priors, recent methods~\cite{2021StyleCLIP, 2021TediGAN-B, 2022PCMFrame} usually use pre-trained text encoders such as \textit{CLIP}~\cite{2021CLIP} and \textit{T5}~\cite{2020T5}, and large-scale generators like StyleGAN~\cite{2019StyleGANv1, 2020StyleGANv2}. In recent years, diffusion models~\cite{2020DDPM, 2020ScoreModel, 2021ImprovedDDPM, 2021ADM, 2022MMDiffusion} have developed rapidly and shown great power in generating images with high quality and diversity. Diffusion-based text-to-image generation methods~\cite{2021GLIDE, 
2022StableDiffusion, 2022DALLE2, 2022Imagen} have achieved impressive results which shocked the world. These methods usually train the model with extremely large datasets like LAION~\cite{2021LAION} and use great computing resources of over thousands of GPU$\cdot$days to fully take advantage of diffusion models.

However, even trained with large datasets and great computing resources, these methods are still limited because texts cannot be detailed enough to describe a highly customized subject. So, it is hard to apply them to synthesize personalized results like images containing a particular dog or a specified car required by the user.

\subsection{Customized Image Generation}

To solve the problem of customized image generation, several methods \cite{2019Obj-GAN, 2021IC-GAN, 2022TI, 2022DreamBooth} have been proposed to provide more control of the generation process. IC-GAN~\cite{2021IC-GAN} uses the representation of certain instances as conditions and trains a corresponding conditional GAN to generate new images of the specified instances.
TI~\cite{2022TI} manages to find a special token representing the specified subject with few-shot data.
DreamBooth~\cite{2022DreamBooth} finetunes a pre-trained text-to-image generation model with few-shot data to push the model to have the ability of generating the costomized subject.
TI and DreamBooth can take both texts and subjects provided by images as conditions to create new images containing the subject because they inject the certain subject into text space. However, since they have to treat every specified subject, their time and computing resource cost are relatively big, which are usually not affordable to most normal users.

\section{Unified Multi-Modal Latent Diffusion}
\label{sec: method}

We aim to generate high-quality images semantically aligned with the input texts while keeping the subjects of the input images.
Our idea is to encode texts and images in a sequence into a unified multi-modal latent embedding and leverage the powerful capabilities of the latent-based diffusion model~\cite{2022StableDiffusion}.
However, it is not easy because of the following non-trivial challenges:
\begin{itemize}
    \item Texts and images are quite different modalities. It is not effortless to encode them into a unified latent space and guarantee that they can both be used to bootstrap the generation process.
    \item User-provided images may contain information irrelevant to the subject, such as complex backgrounds, poses, and lightning. The generation model has a high risk of overfitting to this redundant information and failing to generate new views required by the text.
    \item There is a lack of natural subject-caption-image data pairs for training a joint subject and text conditional image generation framework. 
\end{itemize}

To solve the challenges mentioned above, we propose a UMM-Diffusion model.
First, we make use of text and image encoders pre-trained on large-scale text-image paired data and link them to build a unified multi-modality encoder.
Second, to exclude the influence of irrelevant information in the input image, we propose a novel sampling technique of fusing both results on multi-modal and pure text conditions.
Finally, with a carefully designed training strategy and a dataset of pseudo pairs, we are able to accomplish the novel task of joint subject and text conditional image generation.

In the following, we first review the preliminaries and then introduce our detailed method design.

\subsection{Preliminaries}

\noindent \textbf{Text-to-Image Diffusion Models}.
Diffusion models~\cite{2020DDPM, 2020ScoreModel, 2021ImprovedDDPM} refer to a kind of generative models which first map an arbitrary data distribution to the Gaussian distribution by gradually adding noise, and then generate a sample by gradually denoising a sample from the Gaussian distribution.
The first process is called ``forward process''.
Defining $\mathbf{x}_0$ as a sample from a data distribution $X$, the piecemeal forward process maps $\mathbf{x}_0$ to a sample $\mathbf{x}_T$ from Gaussian distribution by a $T$-step Markovian process:
\begin{equation}
    q(x_t|x_{t-1}) = \mathcal{N}(x_t;\sqrt{1-\beta_t}x_{t-1},\textbf{I}),
\end{equation}
where $\{\beta_t\}$ is a series of hyper-parameters.
The second process is called ``reverse process'', where a noise-prediction model $\bm{\epsilon}_\theta$ is trained to recover the noise of the $t$-step noisy image $\mathbf{x}_t$. The model $\bm{\epsilon}_\theta$ takes both $\mathbf{x}_t$ and $t$ as input.
The training loss is the mean squared error (MSE) between the predicted noise $\bm{\epsilon}_\theta(\mathbf{x}_t, t)$ and the real noise $\bm{\epsilon}$ in $\mathbf{x}_t$.

However, the model $\bm{\epsilon}_\theta$ is unconditional, which means the reverse process is not controllable. To guide the reverse process, an additional condition $c$ (\eg the captions of the images) can be provided to the noise-prediction model $\bm{\epsilon}_\theta$~\cite{2021ADM}. The training loss is then transformed into:
\begin{equation}
\label{eq:mse}
    \mathcal{L} = \mathbbm{E}_{t,\mathbf{x}_0,\bm{\epsilon}}||\bm{\epsilon} - \bm{\epsilon}_\theta(\mathbf{x}_t, c, t)||^2.
\end{equation}

In practice, using only $c$ as the guidance input leads to unsatisfactory results.
To solve this problem, classifier-free guidance~\cite{2022Classifier_Free} has been widely used.
It requires the noise-prediction model $\bm{\epsilon}_\theta$ to be able to process both conditional and unconditional input.
During sampling, the final noise prediction is performed as:
\begin{equation}
\label{eqn: unconditional guidance}
    \Tilde{\bm{\epsilon}}_\theta(\mathbf{x}_t, c, t) = w\bm{\epsilon}_\theta(\mathbf{x}_t, c, t) + (1 - w)\bm{\epsilon}_\theta(\mathbf{x}_t, t),
\end{equation}
where $w$ is the guidance weight. To be more specific, setting $w = 1$ disables classifier-free guidance and $w > 1$ reinforces the effect of guidance input.


\vspace{1mm}
\noindent \textbf{Text Encoding}.
For text-to-image diffusion models, text encoding is vital to both image quality and semantic alignment.
\textit{CLIP}~\cite{2021CLIP} is a widely used encoder for extracting semantic representation from texts and images.
For texts, the words in the texts $T$ are first embedded by a tokenizer $f$. 
Then, the transformer-based text encoder $P$ maps word embedding sequence $w = f(T)$ to a hidden vector sequence $h = P(w)$ with the same length.
Last, $h$ is pooled to the last vector, and the last vector is projected to the final \textit{CLIP Text Embedding}.
However, the last process, which drops the rest of the hidden vector sequence, may lead to insufficient text representations.
There has been CLIP-based text-to-image models~\cite{2022StableDiffusion} using the unpooled hidden vector sequence $h$ instead of the \textit{CLIP Text Embedding} as text condition.

\subsection{Text-and-Image Unified Encoder}
\label{subsec: TIUE}

As stated at the beginning of Sec.~\ref{sec: method}, the first challenge lies in how to encode texts and images into one unified multi-modal latent space.
Existing multi-modal semantic encoders like \textit{CLIP}~~\cite{2021CLIP} encode texts and images into different latent spaces, which are not capable of solving our problem.
From the goal of such a joint encoding, we leverage pretrained text and image independent encoders and build a \textbf{T}ext-and-\textbf{I}mage \textbf{U}nified \textbf{E}ncoder (abbreviated as \textbf{TIUE}), as shown in Fig.~\ref{fig: method}~(a).

\vspace{1mm}
\noindent \textbf{Unified Multi-Modal Encoding}.
We first give an overall formulation of our task.
Denote the input text as $y$, the input images as $\mathbf{x}_s$, and the positions of the words in $y$ which $\mathbf{x}_s$ refers to as $p$.
Our goal is to synthesize a new image $\mathbf{x}$ given a set of $(y, \mathbf{x}_s, p)$ as the input condition.
For example, a sentence ``A \underline{dog} is under a \underline{tower}'' performs as $y$, an image of the dog and another image of the tower perform as $\mathbf{x}_s$, and $p=2,6$ indicates the position of the underlined word ``\underline{dog}'' and ``\underline{tower}''.


Given a set of $(y, \mathbf{x}_s, p)$, TIUE encodes $(y, \mathbf{x}_s, p)$ into a unified vector sequence $h_u$ in a multi-modal latent space which can guide the generation of the whole image $\mathbf{x}$.

As shown in Fig.~\ref{fig: method}~(a), we first resize the input image $\mathbf{x}_s$ and encode it to the \textit{CLIP Image Embedding} with a pretrained \textit{CLIP Image Encoder}.
Then, the \textit{CLIP Image embedding} is projected by a trainable multilayer perceptron (MLP) to a pseudo word embedding so that it can be processed by the \textit{CLIP Text Encoder}.
After that, the pseudo word embedding of the input image $\mathbf{x}_s$ is inserted into the word embedding of the input text $y$ according to the position $p$.
Denoting this rearranged word embedding as $w_r$, we use the pretrained \textit{CLIP Text Encoder} $P$ to map $w_r$ to its corresponding hidden vector sequence $h_r = P(w_r)$.

In order to mitigate the problem of overfitting, we further encode the pure text input $y$ into a hidden vector sequence $h_y$.
Then, we replace the vectors in $h_r$ not at position $p$ (the positions of words that do not correspond to the input image) with the vectors from $h_y$. Finally we obtain a unified vector sequence $h_u = \text{TIUE}(y, \mathbf{x}_s, p)$.

\vspace{1mm}
\noindent \textbf{Training Strategy}.
The supervision of TIUE is non-trivial because it is hard to define the ``unified representation'' from existing data modalities.
To address this problem, we train TIUE along with the whole diffusion process supervised by the MSE loss on noise-prediction in Eqn.~(\ref{eq:mse}):
\begin{equation}
    \mathcal{L} = \mathbbm{E}_{t,\mathbf{x}_0,\bm{\epsilon}}||\bm{\epsilon} - \bm{\epsilon}_\theta(\mathbf{x}_t, \text{TIUE}(y, \mathbf{x}_s, p), t)||^2.
\end{equation}
We freeze \textit{CLIP Encoder}s and only train the MLP projecting image embeddings into pseudo word embeddings.
Data preparation for this training is introduced in Sec.~\ref{subsec: dataset and initialize}.

\subsection{Fusing Sampling Technique}
\label{subsec: fuse sample}

TIUE takes the whole image $\mathbf{x}_s$ as input, including both the subject we want the model to condition on and the irrelevant background.
Such behavior leads to the second challenge of overfitting, as stated at the beginning of Sec.~\ref{sec: method}.
We notice that the pure text input $y$ is not influenced by the undesired parts of $\mathbf{x}_s$.
So, we propose to use $y$ as an assistance to solve the overfitting problem by applying both the multi-modal guidance and pure text guidance and fusing their results during the sampling process of 
the diffusion model. The overall design is illustrated in Fig.~\ref{fig: method}~(b).

To be more specific, on the $t$-th step of the diffusion process, the noise-prediction model $\bm{\epsilon}_\theta$ originally predicts the noise $\bm{\epsilon}$ contained in the noisy image $\mathbf{x}_t$ with the unified multi-modal condition $h_u$ and timestep $t$ as extra input:
\begin{equation}
    \hat{\bm{\epsilon}}_u = \bm{\epsilon}_\theta(\mathbf{x}_t, h_u, t).
\end{equation}
The problem is that $\hat{\bm{\epsilon}}_u$ suffers from the influence of irrelevant context encoded in $h_u$.
Meanwhile, the prediction conditioned on the pure text $h_y$ is not affected by $h_u$:
\begin{equation}
    \hat{\bm{\epsilon}}_y = \bm{\epsilon}_\theta(\mathbf{x}_t, h_y, t),
\end{equation}
The limitation of $\hat{\bm{\epsilon}}_y$ is that it does not contain the semantic of the specified subject in the input image $\mathbf{x}_s$ and only contains general subjects described by the input text $y$.
For example, taking ``A \underline{dog} running on the beach'' and an image of a particular dog as input, using only $\hat{\bm{\epsilon}}_u$ would transfer the background of the dog image and using only $\hat{\bm{\epsilon}}_y$ would lead to an ordinary dog.

Considering the effect of both of $\hat{\bm{\epsilon}}_u$ and $\hat{\bm{\epsilon}}_y$, we propose to use both of them and fuse their noise-prediction outputs in the sampling process:
\begin{equation}
\label{eqn: fusing sample}
    \hat{\bm{\epsilon}}_f = \alpha\hat{\bm{\epsilon}}_u + (1 - \alpha)\hat{\bm{\epsilon}}_y.
\end{equation}
The fuse ratio $\alpha \in [0, 1]$ leads to a trade-off between overfitting and semantic aligning.
When the background of the input image is complicated and conflicts with the text description, $\alpha$ should be lower to mitigate the negative effect of the background, and vice versa.
In the corner cases, $\alpha = 0$ represents pure text-to-image generation, and $\alpha = 1$ disables the fusing sampling technique.

Combining the fusing sampling technique (Eqn.~(\ref{eqn: fusing sample})) and classifier-free guidance (Eqn.~(\ref{eqn: unconditional guidance})), in practice we estimates:
\begin{equation}
\label{eqn: total sample}
\begin{aligned}
    \Tilde{\bm{\epsilon}}_\theta = \alpha w\hat{\bm{\epsilon}}_u + (1 - \alpha)w\hat{\bm{\epsilon}}_y + (1 - w)\bm{\epsilon}_\theta(\mathbf{x}_t, t).    
\end{aligned}
\end{equation}
where $\Tilde{\bm{\epsilon}}_\theta$ in conditioned on the complete input set $(y, \mathbf{x}_s, p)$.

The sampling technique in Eqn.~(\ref{eqn: total sample}) requires that the model support both unconditional input, unified multi-modal condition input $h_u$, and pure text condition input $h_y$.
Therefore, we train the model by randomly giving one of these three conditions.
The probability of each condition is set to 0.1, 0.54=0.9$\times$0.6, and 0.36=0.9$\times$0.4, respectively, to ensure that the model achieves the requirement of fusing different guidance.

\begin{figure*}[htbp]
    \centering
    \includegraphics[width=\linewidth]{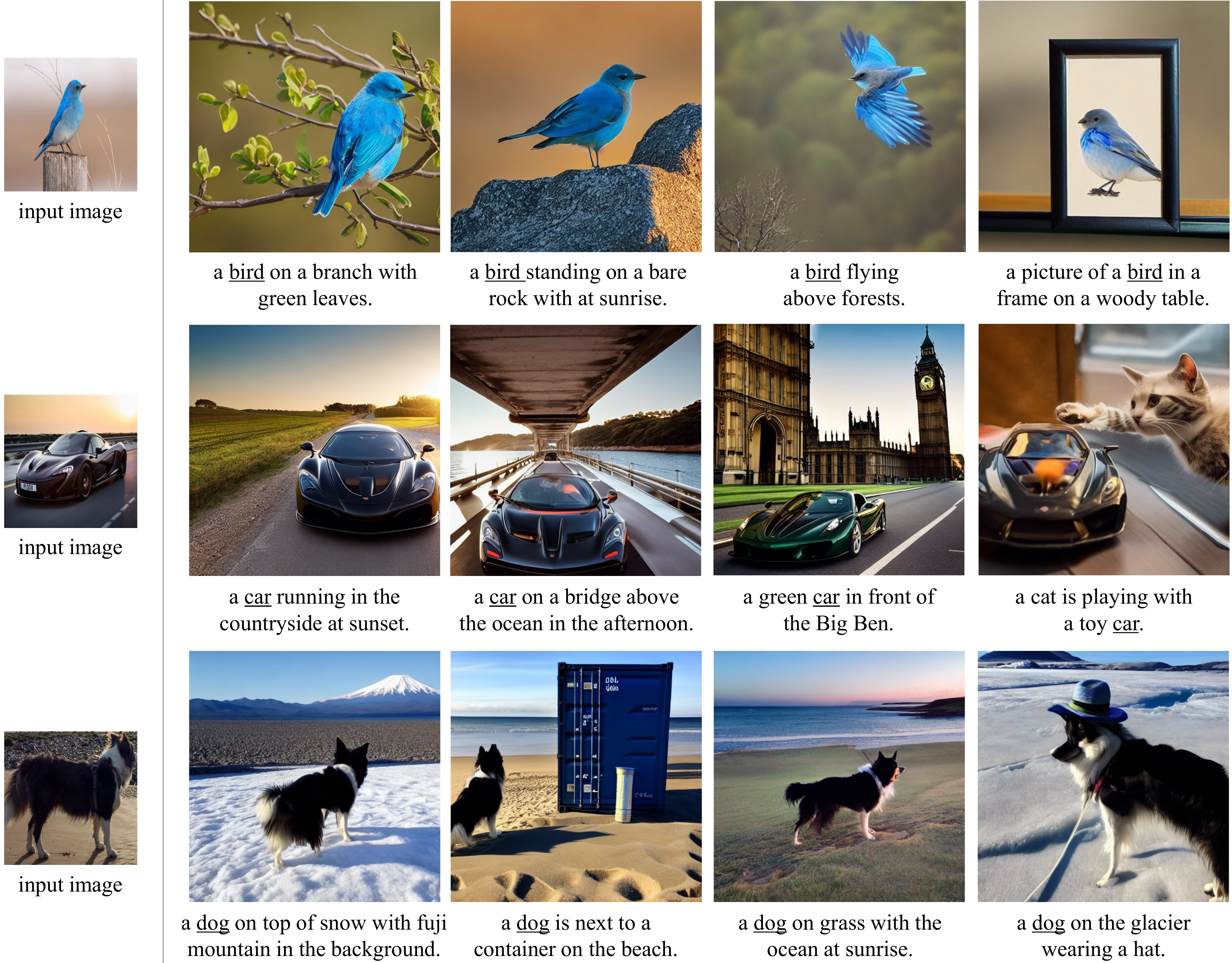}
    \caption{Results of multi-modality latent embedding guided image generation. Our method is able to create new scenes corresponding to the input texts with the specified subject provided by input images. For one subject, the proposed method only requires one input image and \textbf{does not need any finetuning}.}
    \label{fig: base results}
\end{figure*}

\subsection{Dataset Building and Model Initialization}
\label{subsec: dataset and initialize}

To train the model, we need a dataset containing sets of $(\mathbf{x}, y, \mathbf{x}_s, p)$.
We leverage a widely used text-image dataset, LAION-400M \cite{2021LAION}. 
Given a data sample $(\mathbf{x}, y)$ in LAION-400M, where $\mathbf{x}$ is the image and $y$ is the corresponding caption, we leverage an object detection model~\cite{2015FastRCNN} to automatically crop sub-images $\mathbf{x}_s$ containing particular objects.
Then, we retrieve the labels of $\mathbf{x}_s$ in the caption $y$.
If the label is found, we record the position $p$ of the word.
In this way, we obtain $(\mathbf{x}, y, \mathbf{x}_s, p)$ sets.

To ensure data quality, we discard the following cases:
\begin{itemize}
    \item Too small bounding boxes occluding less than 10\% of the whole image. These cases lead to ineffective image guidance which confuses the generation model.
    \item Multiple bounding boxes corresponding to the same label in one image. These cases lead to ambiguous word references.
    \item Low-resolution images (less than 512$\times$512). These cases are low-quality training data.
\end{itemize}
Under these rules, we filter LAION-400M to a subset containing $\sim$1.8M training sets.
Such a scale of training data is not enough to train our model from scratch.
Existing text-to-image diffusion models are mostly trained with hundreds of millions of text-image data pairs.
Multi-modality guided image generation can be even more difficult than pure text-to-image generation. 

To solve the problem of data limitation and save the effort of training, we utilize a pre-trained text-to-image generation model to initialize the parameters of our model.
Specifically, we leverage Stable Diffusion v1-5~\cite{2022StableDiffusion} and replace its unpooled \textit{CLIP Text Encoder} with the proposed TIUE.
The MLP in TIUE is randomly initialized.
The training process contains two phases. First, the noise-prediction model is frozen to serve as the supervision of training TIUE.
Then we optimize the whole model.
The pre-trained text-to-image model already has the ability to process the semantics contained in texts.
Therefore, phase one is the core of training, \ie, learning a unified multi-modal latent space, and phase two further improves the generation performance.


\section{Experiments}


\subsection{Implementation Details}
\label{subsec: implementation}


We first train only TIUE for 200k iterations and then train the whole model for another 200k iterations. The batch size is 192 and the learning rate is 1e-5 for both two phases. We use an EMA rate of 0.9999 to stabilize the training process. During inference, the fuse ratio $\alpha$ is set to 0.5.

To the best of our knowledge, we are the first to design a unified encoder that takes joint subjects (provided by images) and texts as conditions for image generation.
Due to a lack of other works under the same task, we compare the results of our method with text-to-image baselines and few-shot finetuning methods on similar tasks. All results are sampled by DDIM~\cite{2020DDIM} with 50 steps.


\begin{figure}
    \centering
    \includegraphics[width=\linewidth]{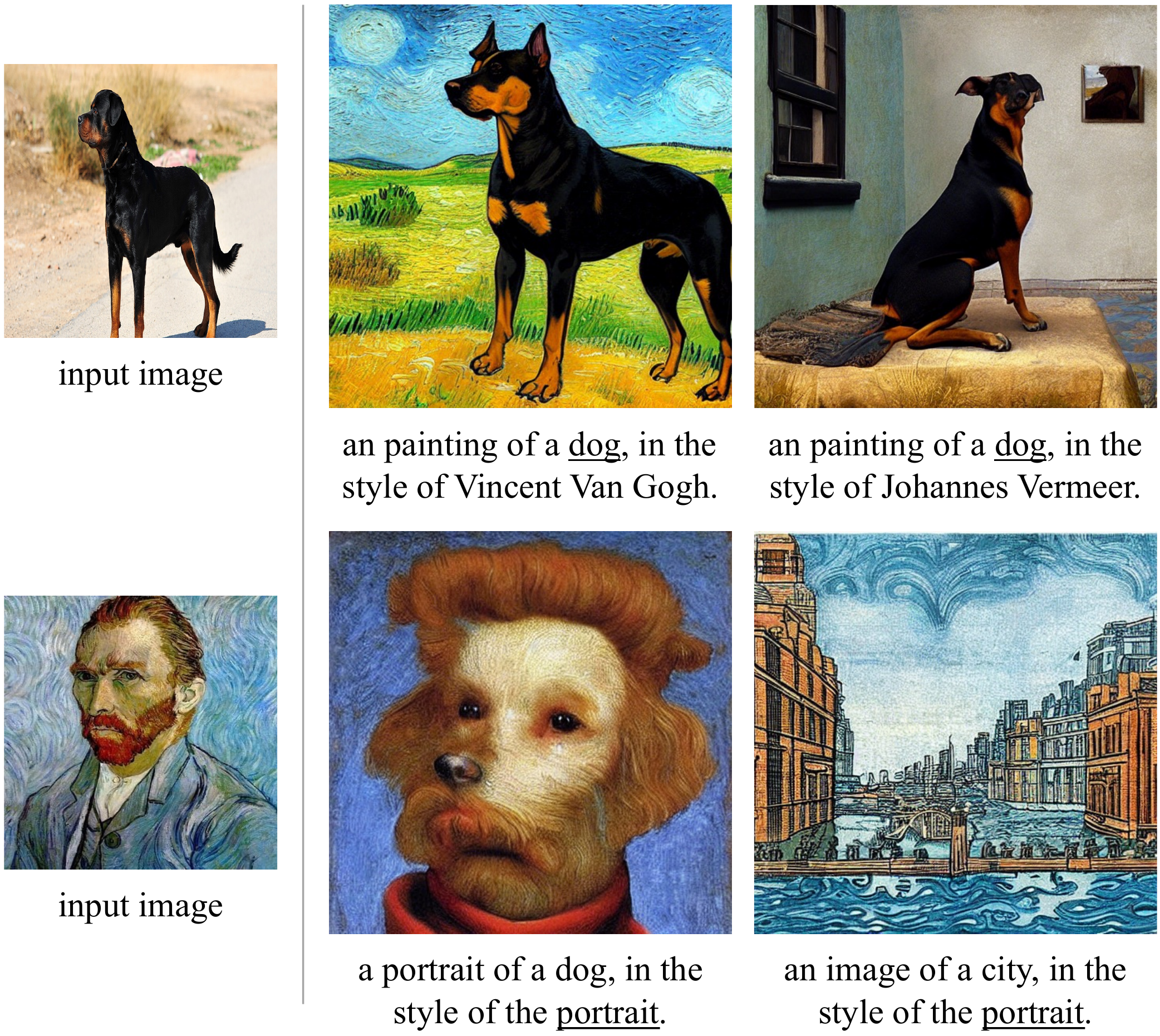}
    \caption{Stylized results. First line presents results with styles provided by texts and the second line shows results with styles provided by images.}
    \label{fig: style results}
\end{figure}

\subsection{Applications}


\noindent \textbf{Multi-Modality Guided Image Generation.}
Given a set of $(y, \mathbf{x}_s, p)$, our method can generate images that are described by $y$ and contain the object in $\mathbf{x}_s$ as shown in Fig.~\ref{fig: base results}.
Benefiting from diffusion models, our model can generate diverse novel views of the target subject, including new backgrounds, positions, lighting conditions, and status. 

\vspace{1mm}
\noindent \textbf{Image Synthesis with Customized Style.}
Since our model supports joint subject and text conditions, the style of the result can be either assigned by the input text $y$ or the input image $\mathbf{x}_s$. We show several results in both ways of assigning styles in Fig.~\ref{fig: style results}.
Note that in the second row, the input prompt  
``in the style of'' indicates that the model should only extract the style of $\mathbf{x}_s$ and omit its content. Results show that the proposed model can generate images in the style of $\mathbf{x}_s$ and not be affected by the semantics of the content.
It demonstrates that our model is able to disentangle the information of $y$ and $\mathbf{x}_s$.


\begin{figure}
    \centering
    \includegraphics[width=\linewidth]{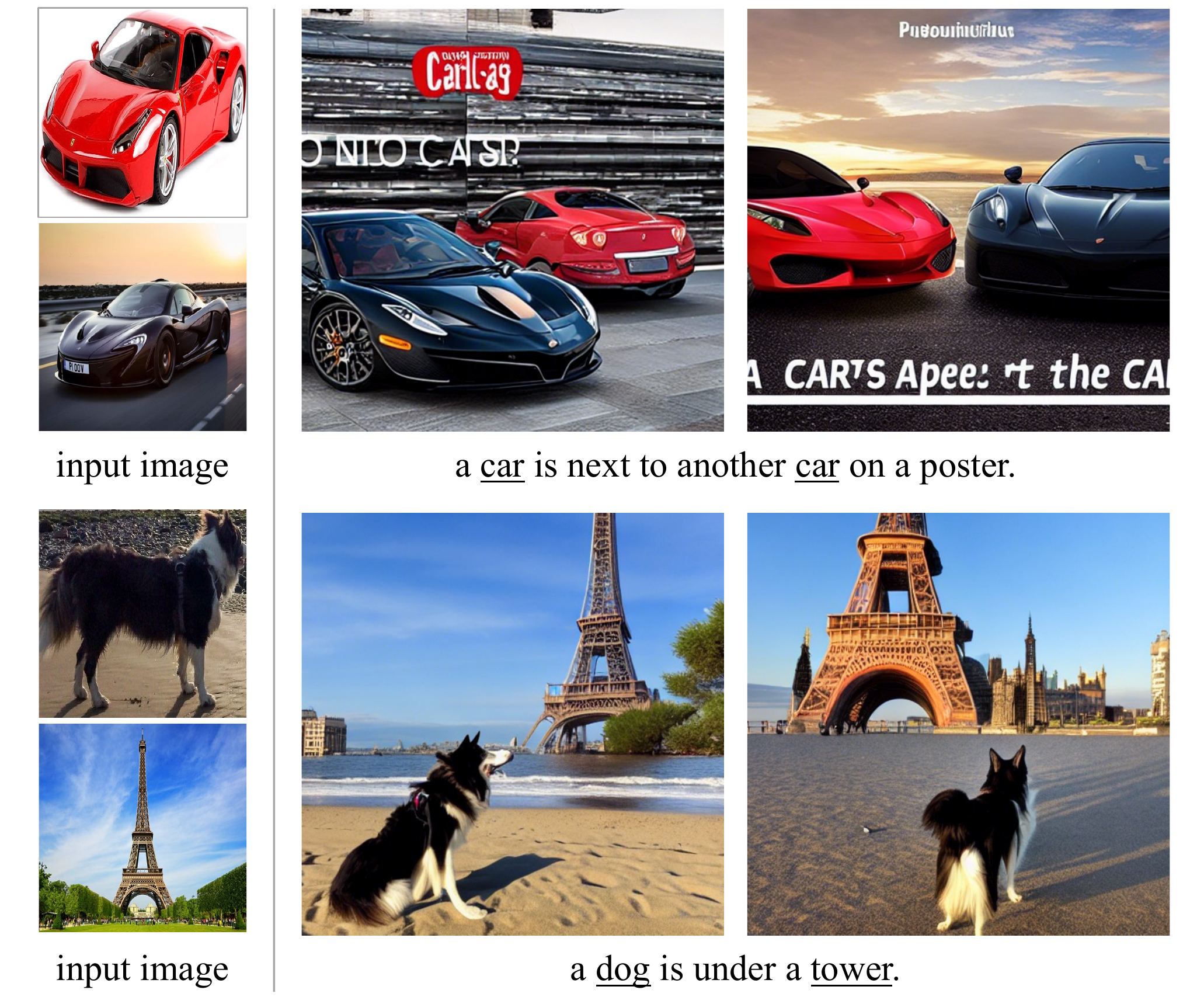}
    \caption{Results synthesised by multiple image guidance in one input set.}
    \label{fig: multiple input results}
\end{figure}

\vspace{1mm}
\noindent \textbf{Multiple Image Guidance in One Input Set.}
Our model allows the users to provide more than one $\mathbf{x}_s$ in the input set. In other words, we can specify multiple objects in the input sentence $y$.
We show several such results in Fig.~\ref{fig: multiple input results}. The results show that our method is able to recover the visual features of multiple input images and organize them harmoniously corresponding to the input texts, \eg consistent shadows and appropriate sizes of different subjects. 


\begin{figure*}
    \centering
    \includegraphics[width=\textwidth]{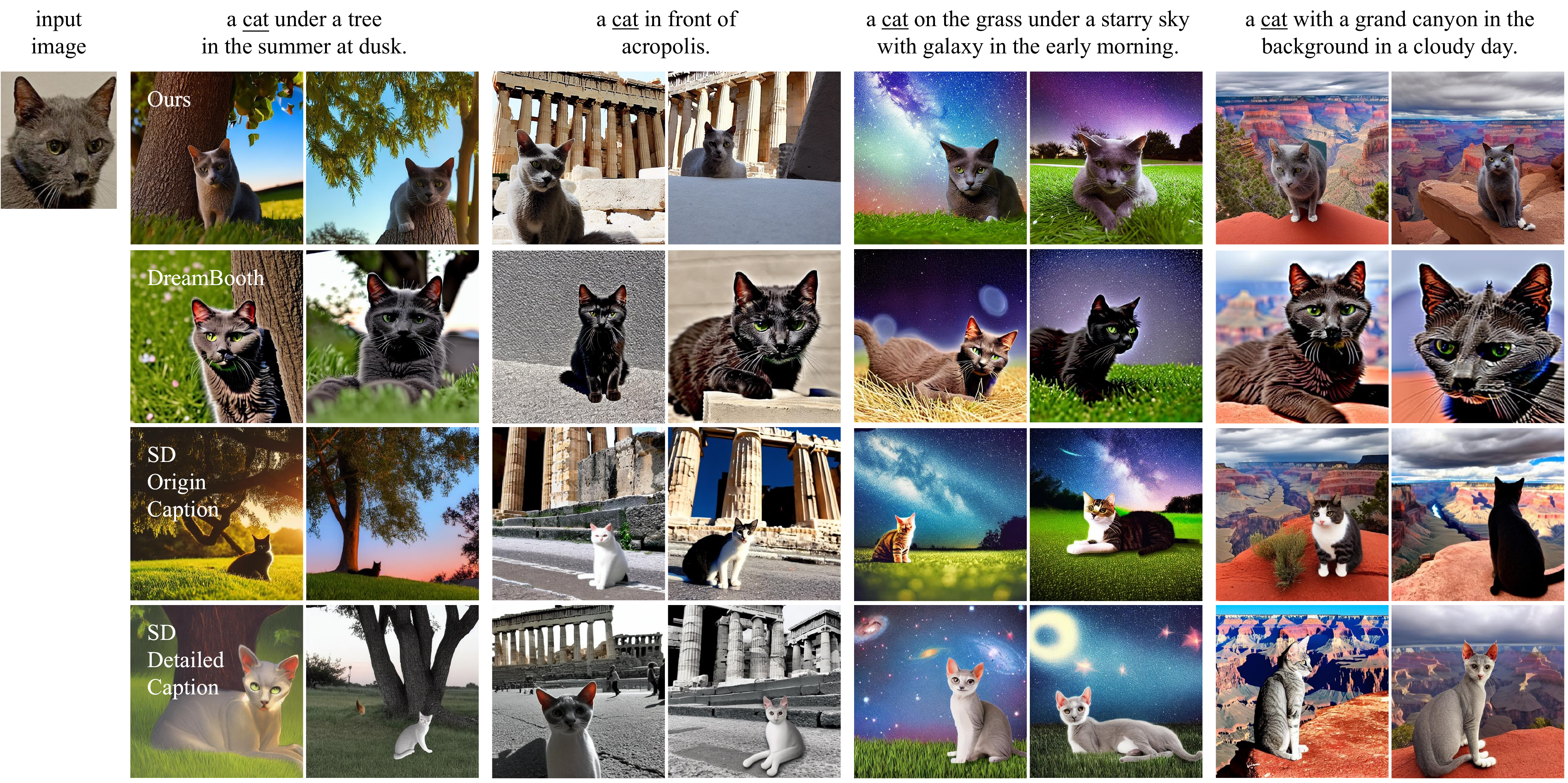}
    \caption{Comparisons between our method and several baselines. Our method can get comparable or better results comparing with DreamBooth \cite{2022DreamBooth}, a few-shot finetuing method. ``SD'' denotes ``Stable Diffusion''. ``Detailed Caption'' built on SD replaces ``cat'' with ``gray, thin cat with pointed ears'', which still generates unsatisfying results.  \textbf{[Zoom in for best view]}}
    \label{fig: baseline results}
\end{figure*}

\begin{figure}
    \centering
    \includegraphics[width=\linewidth]{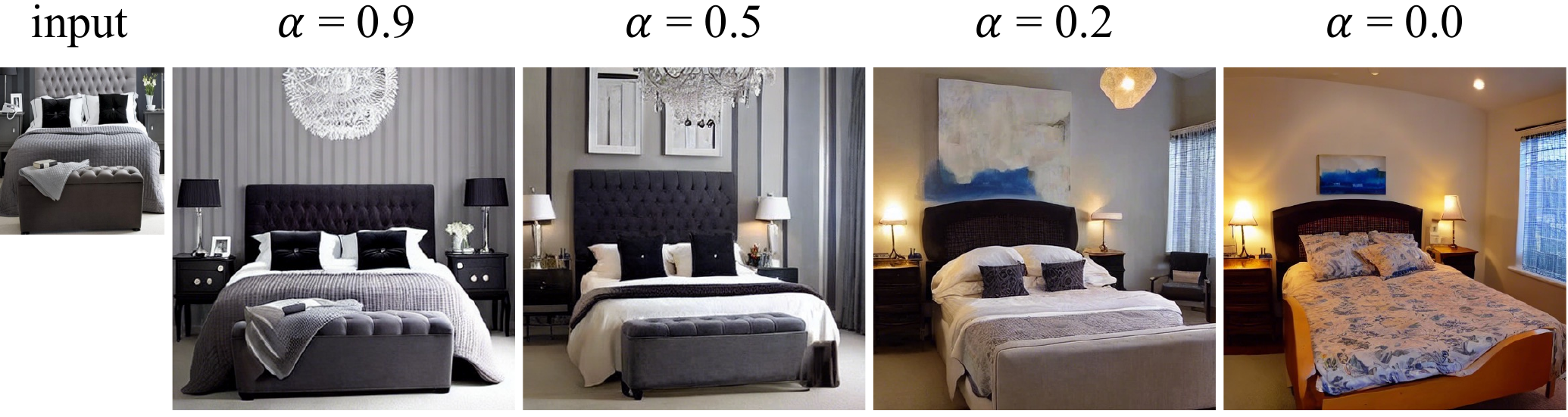}
    \caption{Ablation on the fuse ratio $\alpha$. Prompt: a big \underline{bed} in a bedroom with paintings on the wall and lamps beside it.
}
    \label{fig: ablation results}
\end{figure}

\subsection{Comparisons}
\label{subsec: comparison}
As we addressed before, we are the first to build a unified encoder to achieve the goal of multi-modality image generation.
To demonstrate the superiority of the proposed method, we design several text-to-image baselines to compare, which take original texts $y$ or more detailed descriptions as input.
Furthermore, we compare our method to a state-of-the-art few-shot finetuning-based method, DreamBooth \cite{2022DreamBooth}. Note that DreamBooth takes dozens of minutes to tune the model on GPUs.
In comparison, our model has no online training cost, which is more convenient in real applications.

The results in Fig.~\ref{fig: baseline results} prove that we can get better results without the time-costing finetuning process. When only given one input image with low resolution, DreamBooth may overfit and fail to generate a correct scene indicated by input text (\eg the second input text, ``in front of acropolis''). Pure text-to-image methods cannot be customized enough even with a detailed caption.



\subsection{Ablation Studies}

We conduct an experiment to demonstrate the ability of the fusing sampling technique proposed in Sec.~\ref{subsec: fuse sample} and show the trade-off on the choice of $\alpha$.
The results are shown in Fig~\ref{fig: ablation results}.
It can be seen that, 
when alpha is over small, the generated images are highly-aligned with the input text but lack the visual features of the provided subject (\eg the color and texture of pillows distort at $\alpha=0.2$ or $\alpha=0.0$). When alpha is over large, the generated images can keep details of the subject while do not align with the text description (\eg the result misses the requirement ``paintings on the wall'' in the input text at $\alpha=0.9$).
$\alpha=0.5$ usually leads to a better trade-off between input texts and images.
So we set $\alpha=0.5$ to generate the results in the paper.

\subsection{Limitations}
\label{subsec: limitation}

\begin{figure}
    \centering
    \includegraphics[width=0.98\linewidth]{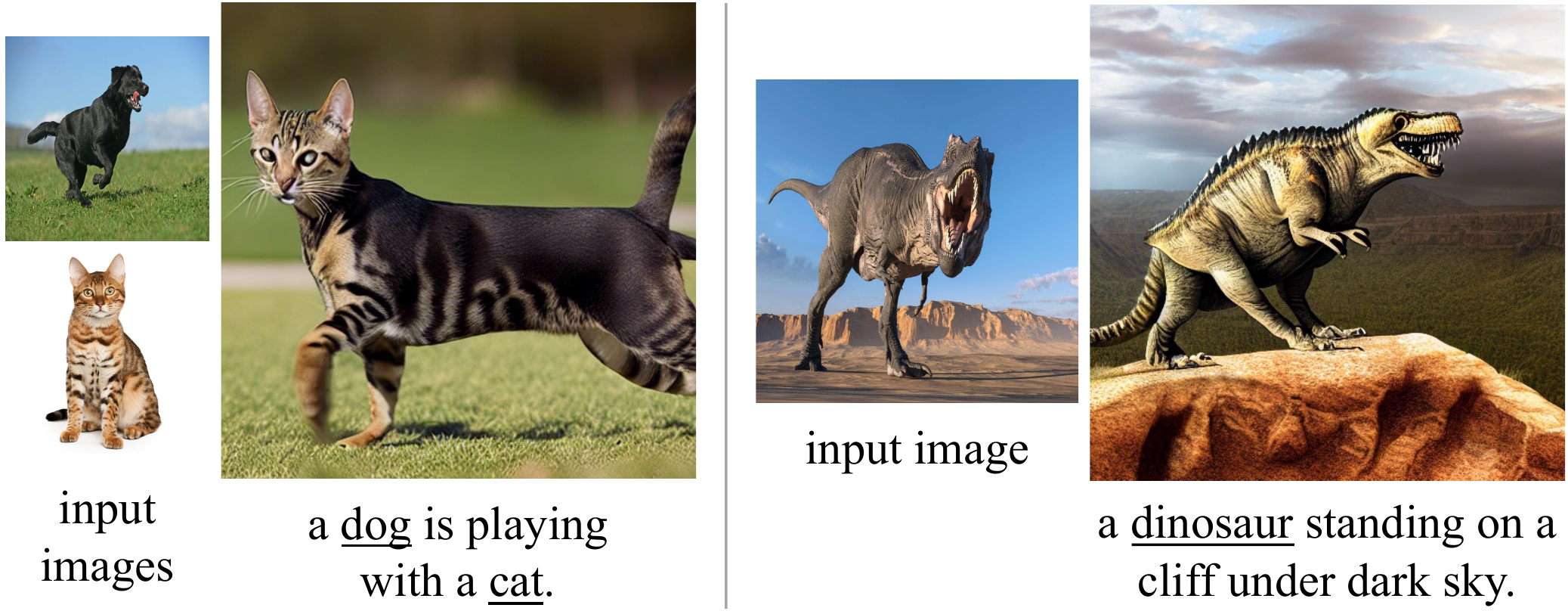}
    \caption{Failure cases. Multiple image guidance may lead to feature composition. Rare or highly-factitious subjects leads to distorted or wrong visual features.}
    \label{fig: failure results}
\end{figure}

There are two main drawbacks of our method.
First, when there are multiple image guidance $\mathbf{x}_s$, our model might mix the features of different subjects and generates a fused one as shown in the left of Fig.~\ref{fig: failure results}. That is because \textit{CLIP Text Encoder}, the text encoder we use, performs poorly on multi-object decomposition, which has been explored in previous works \cite{2022DALLE2, 2022Imagen, 2022StructuredDiffusion}.
Second, when facing rare subjects (\eg dinosaurs) or highly-factitious subjects, our method may distort or lose some details of the target subject as shown in the right of Fig.~\ref{fig: failure results}.

\section{Conclusion}

In this paper, we propose a UMM-Diffuision, a framework for joint subject and text conditional image generation with a corresponding encoder to encode such multi-modal data into one unified latent space.
The encoder leverages the power of \textit{CLIP Encoder}s to extract the semantics of texts and images.
To mitigate the problem of overfitting on the irrelevant part of input images, we present a fusing sampling technique which uses both multi-modal and pure text guidance.
The model is initialized by a pre-trained text-to-image generation model.
The initialization solves the problem of limited data and saves the effort of training.
Diverse experiment results demonstrate the efficiency of our method.

{\small
\bibliographystyle{ieee_fullname}
\bibliography{used_ref}
}

\newpage

\appendix

\section*{Supplementary Materials}

\section{Further Implementation Details}

Basic implementation details have been addressed in the Sec.~\ref{subsec: implementation} of main paper. We present more details here.
We train our model on 32 NVIDIA V100 32G GPUs for about 7 days with full precision. We use Adam \cite{2014Adam} optimizer during training. During inference, the scale factor of classifer-free guidance \cite{2022Classifier_Free} is set to 7.5 following Stable Diffusion \cite{2022StableDiffusion}. The \textit{CLIP Encoder}s \cite{2021CLIP} we used are ViT-L/14 based. The noise-prediction model architecture is the same to the origin model of Stable Diffusion \cite{2022StableDiffusion} v1-5. 

The authors of DreamBooth \cite{2022DreamBooth} do not release the official Imagen-based \cite{2022Imagen} code, so we use an unofficial implementation \cite{2022DreamBoothOnline} based on Stable Diffusion \cite{2022StableDiffusion}.

\section{Advantages of Our Method Comparing with Few-Shot Finetuning Methods}
\label{sec: advantage}

\noindent\textbf{Less Running Time and Lower Requirement of Computing Resources.} Our method does not require any finetuning on an input set. Comparing the few-shot finetuning based methods \cite{2022DreamBooth, 2022TI}, our method requires much less time and computing resource of only inference. The quantitative comparison on preparing time (does not include the inference time) is shown in Tab.~\ref{tab: time comparison}. It should be noticed that NVIDIA V100 32G GPUs are not affordable to most normal users, so these few-shot finetuning based methods are actually not practical enough.

\begin{table}[htbp]
    \centering
    \begin{tabular}{ccc}
    \toprule
         & Optimize Steps & Time Cost 
       \\
       \midrule
       TI \cite{2022TI} & 5000 & $\sim 3000$
       \\
       \midrule
       DreamBooth \cite{2022DreamBooth} & 500-800 & $\sim 1000$
       \\
       \midrule
       Ours & 0 & 0
       \\
       \bottomrule
    \end{tabular}
    \caption{The comparison on preparing time (without inference time). The time cost is a rough value performed on one single NVIDIA V100 32G GPU. Our method does not need any preparing for one input set $(y, \mathbf{x}_s, p)$. The number of DreamBooth is performed by the unofficial implementation \cite{2022DreamBoothOnline} based on Stable Diffusion \cite{2022StableDiffusion} rather than Imagen \cite{2022Imagen}, while Stable Diffusion takes much less time for finetuning.}
    \label{tab: time comparison}
\end{table}

\begin{figure}
    \centering
    \includegraphics[width=0.9\linewidth]{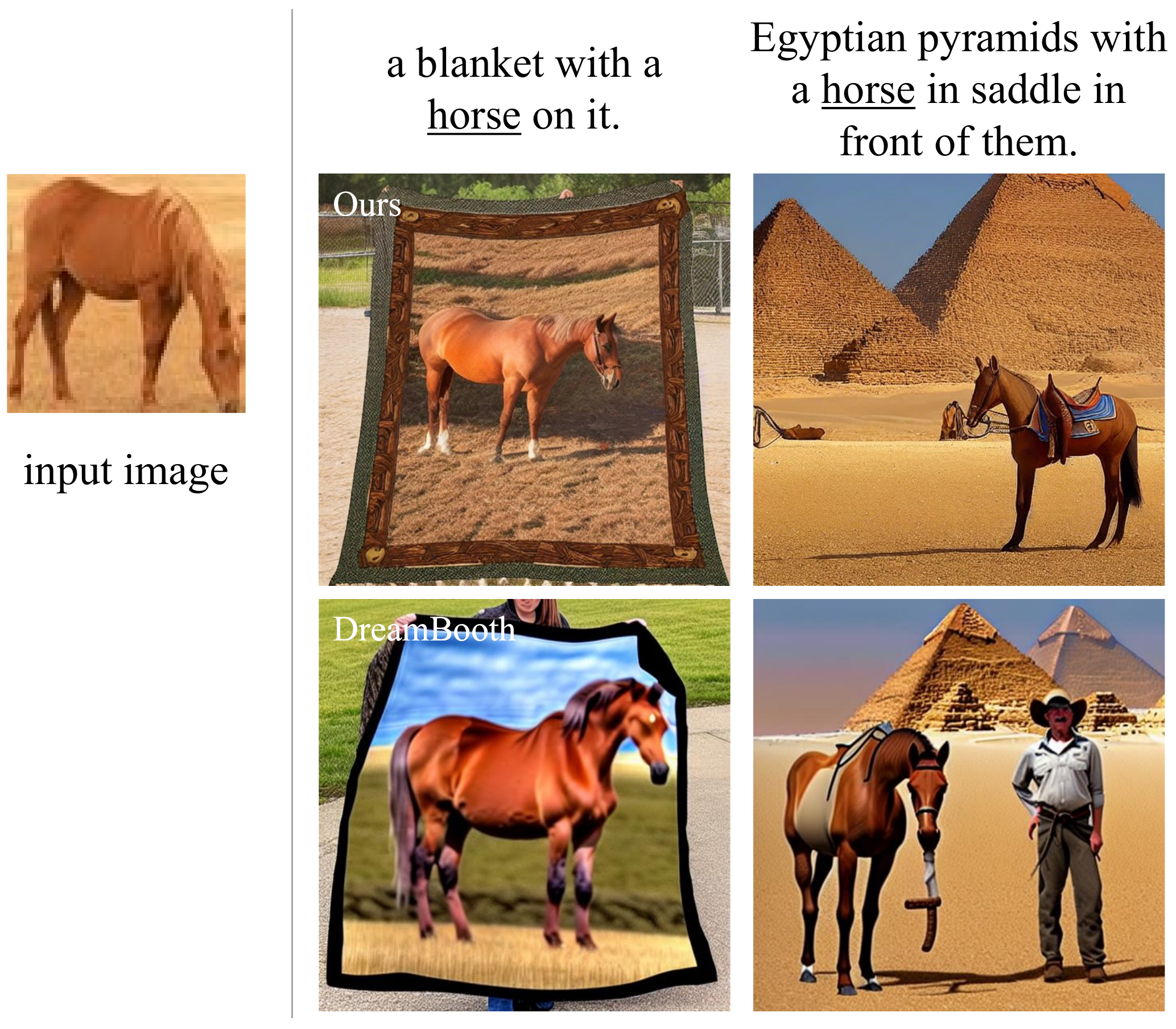}
    \caption{The comparison of results with a low-quality image provided between the proposed method and DreamBooth \cite{2022DreamBooth}, a few-shot finetuning based method. It can be seen that the few-shot finetuning based method generates more blurry results than ours. \textbf{[Zoom in for best view]}}
    \label{fig: more comparison}
\end{figure}

\begin{figure*}[htbp]
    \centering
    \includegraphics[width=\linewidth]{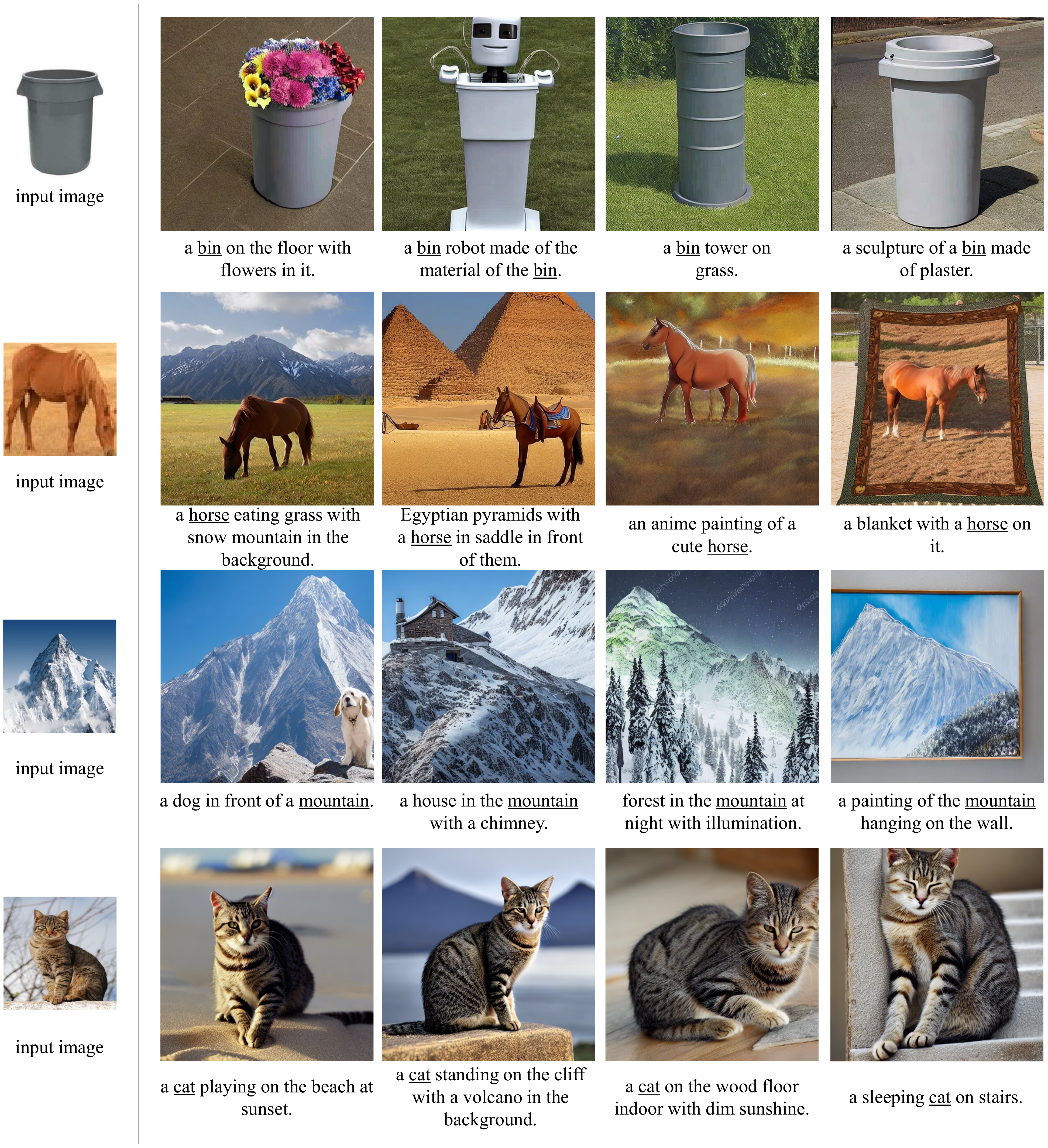}
    \caption{Multi-modality guided image generation results of our method. The subjects provided by input images can be used in several ways. Our method is tolerant to the quality of input images (\eg in the second row, the image of the horse has low resolution, while our method still can generate high quality images corresponding to the inputs). For few-shot finetuning methods, low quality of provided images would lead to bad performance because they use the provided images as the outputs of the pre-trained generation model while the outputs have high quality.}
    \label{fig: more multi-modal guided results}
\end{figure*}

\begin{figure*}[htbp]
    \centering
    \includegraphics[width=\linewidth]{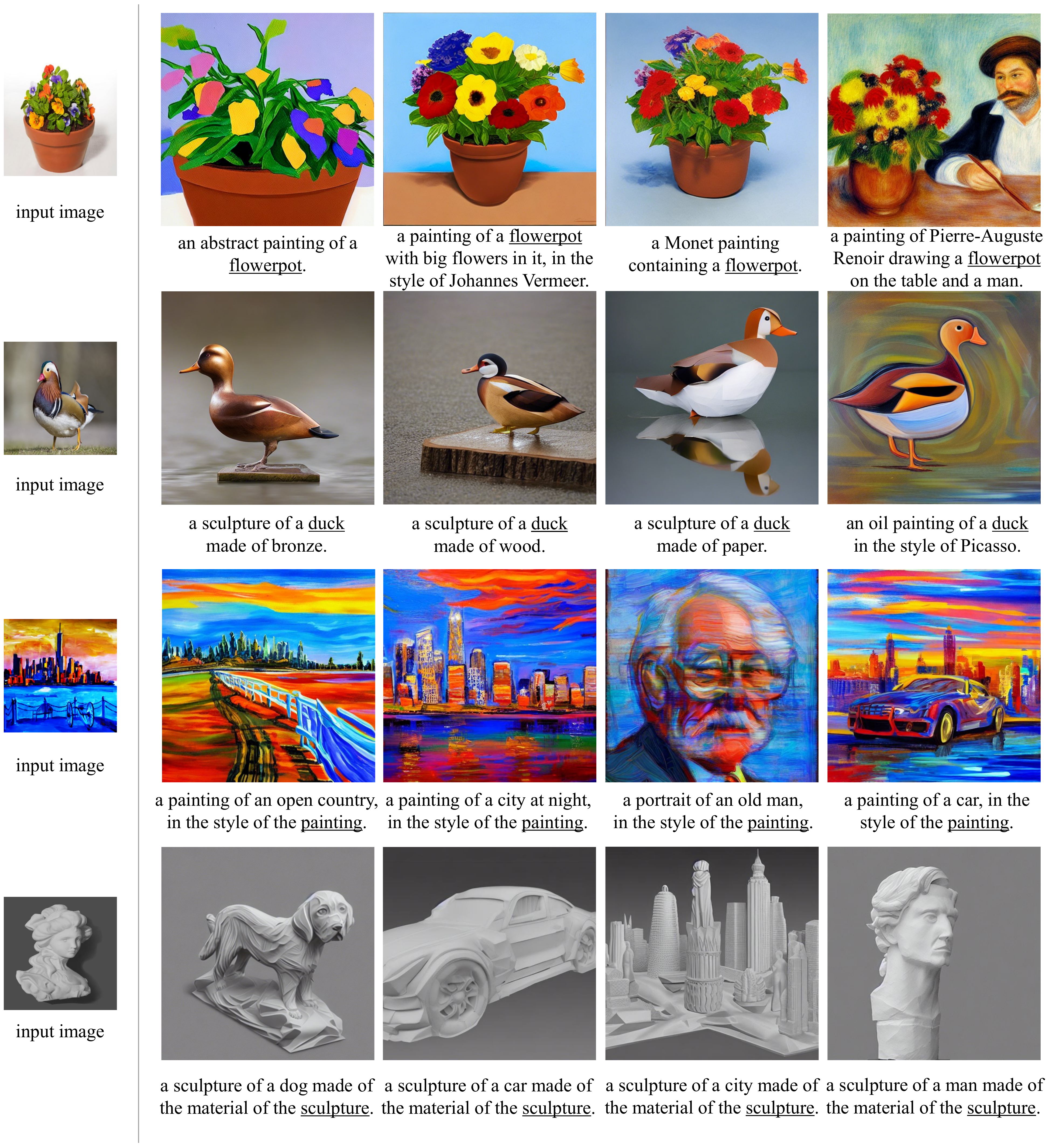}
    \caption{Stylized results of our method. ``Stylization'' means both style of paintings and materials of sculptures. The customized styles can be provided by both texts (the top two rows) and subjects (the bottom two rows). Our method is able to extract contents and styles from the provided images separately, performing well on decomposition of content and style especially in the results of bottom two rows. In these results, our method generates images of styles provided by subjects excluding the contents of subjects.}
    \label{fig: more stylized results}
\end{figure*}

\noindent\textbf{Tolerance to Low-Quality Input Images.} Different from few-shot finetuning based methods \cite{2022DreamBooth, 2022TI}, our method is able to take low-quality images as inputs and generate high quality images as results. Our method encodes input images to compact representation embeddings, which means the quality of these images will not affect a lot on the encoded embeddings. Meanwhile, few-shot finetuning based methods use provided images as outputs of pre-trained generation models to finetune them. However, the outputs of pre-trained generation models are high-quality images. If the quality of provided images are lower than the outputs, they would generate blurry images because they use low-quality images to finetune the model. The comparison shown in Fig.~\ref{fig: more comparison}. Besides, few-shot finetuning based methods also need more than one image to perform better, unless they might overfit on one single image.

\section{Further Experiments}

\noindent\textbf{Multi-Modality Guided Image Generation.}
We show several results on more diverse subjects and texts in Fig.~\ref{fig: more multi-modal guided results} to demonstrate the superiority of our method. As we discussed before in Sec.~\ref{sec: advantage}, different from few-shot finetuning based methods, our method can take low-resolution images as input images and generate high-quality images as results as the second row in Fig.~\ref{fig: more multi-modal guided results}. The subjects provided by images can be used as objects, backgrounds or materials.

\vspace{1mm}
\noindent\textbf{Image Synthesis with Customized Style.}
Due to the limitation of the length of the main paper, we show more stylized results in Fig.~\ref{fig: more stylized results}. ``Stylization'' does not only mean ``paintings in a certain style'', but also mean ``objects made of a certain material''. It can be seen that our method is able to disentangle contents and styles of input images and use them to create new images according to the input prompts.

\begin{figure}
    \centering
    \includegraphics[width=\linewidth]{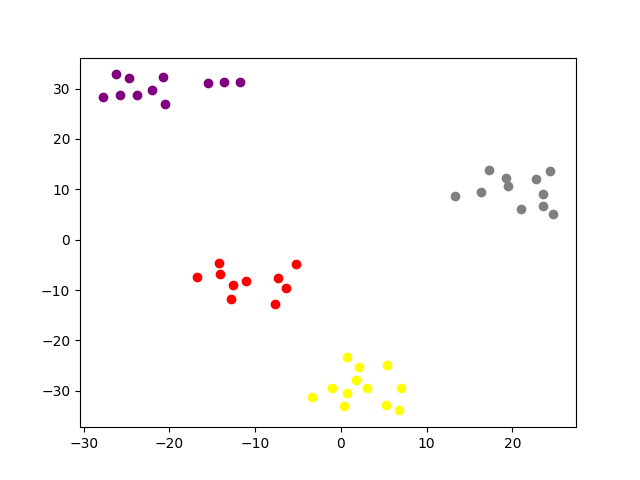}
    \caption{The visualization of pseudo word embeddings from four different classes via t-SNE. Embeddings of ``dog'' are in red, ``cat'' are in gray, ``horse'' are in yellow and ``car'' are in purple. The division is clear, proving the proposed TIUE can extract semantic representations of different classes of images.}
    \label{fig: visualization}
\end{figure}

\vspace{1mm}
\noindent\textbf{Visualization of Pseudo Word Embeddings.} As we addressed in the main paper, the proposed TIUE encodes images to pseudo word embeddings so that they can be encoded into one unified multi-modal latent space along with texts. In other words, TIUE is able to extract similar embeddings from images of one class and extract significantly different embeddings from images of different classes. To prove this, we choose 44 images of four classes including ``dog'', ``cat'', ``horse'', and ``car'' and extract their pseudo word embeddings through TIUE. Then we visualize them through t-SNE \cite{2008tSNE} in Fig.~\ref{fig: visualization}. It can be seen that the pseudo word embeddings from different classes are divided into different clusters, demonstrating our claim.

\end{document}